\documentclass{article}

\usepackage[nonatbib,preprint]{neurips_2019}

\usepackage[utf8]{inputenc} 
\usepackage[T1]{fontenc}    
\usepackage{hyperref}       
\usepackage{url}            
\usepackage{booktabs}       
\usepackage{amsfonts}       
\usepackage{nicefrac}       
\usepackage{microtype}      

\usepackage{graphicx}
\usepackage{subfigure}

\title{Understanding Human Judgments of Causality}

\author{%
  Masahiro Kazama$^1$, Yoshihiko Suhara$^{2}$, Andrey Bogomolov$^{2}$, and  Alex `Sandy' Pentland$^{2}$ \\
  $^1$Habitech Inc., 
  $^2$Massachusetts Institute of Technology
  \\
  \texttt{kazama.masa@gmail.com},  \texttt{\{suhara,abogomol,pentland\}@mit.edu}
}

\begin{document}

\maketitle

\begin{abstract}
 Discriminating between causality and correlation is a major problem in machine learning, and theoretical tools for determining causality are still being developed. However, people commonly make causality judgments and are often correct, even in unfamiliar domains. What are humans doing to make these judgments? This paper examines differences in human experts' and non-experts' ability to attribute causality by comparing their performances to those of machine-learning algorithms. We collected human judgments by using Amazon Mechanical Turk (MTurk) and then divided the human subjects into two groups: experts and non-experts. We also prepared expert and non-expert machine algorithms based on different training of convolutional neural network (CNN) models. The results showed that human experts' judgments were similar to those made by an ``expert'' CNN model trained on a large number of examples from the target domain. The human non-experts' judgments resembled the prediction outputs of the CNN model that was trained on only the small number of examples used during the MTurk instruction. We also analyzed the differences between the expert and non-expert machine algorithms based on their neural representations to evaluate the performances, providing insight into the human experts' and non-experts' cognitive abilities.
\end{abstract}

\section{Introduction}
Understanding human cognitive abilities is essential from not only a scientific perspective but also for many practical applications. A deep understanding of human cognitive abilities is helpful in the collaboration between humans and machines because it allows us to understand the limits of human workers' abilities. The cognitive abilities of human workers are diverse due to their diverse experiences. For any task, human workers can be divided into {\it experts} and {\it non-experts}, and we can utilize experts' knowledge of a specific task to support non-experts' understanding of the task if we can differentiate experts and non-experts. This {\it Extended Intelligence} \cite{ExtendedInte2016} approach can potentially enhance human workers' performances through computational algorithms that are compatible with human cognitive representations.

This paper focuses on human cognitive abilities to perform a {\it cause-effect attribution} task \cite{Mooij:2016uf}. The task was to attribute causality between two variables after making training observations of the two variables. We chose the task because causality judgments are important and yet unfamiliar to most people; thus, the task was appropriate for understanding the differences between human experts' and non-experts' cognitive abilities.

\begin{figure}[t] \label{fig:paper_summary}
    \centering
    \includegraphics[width=0.8\textwidth]{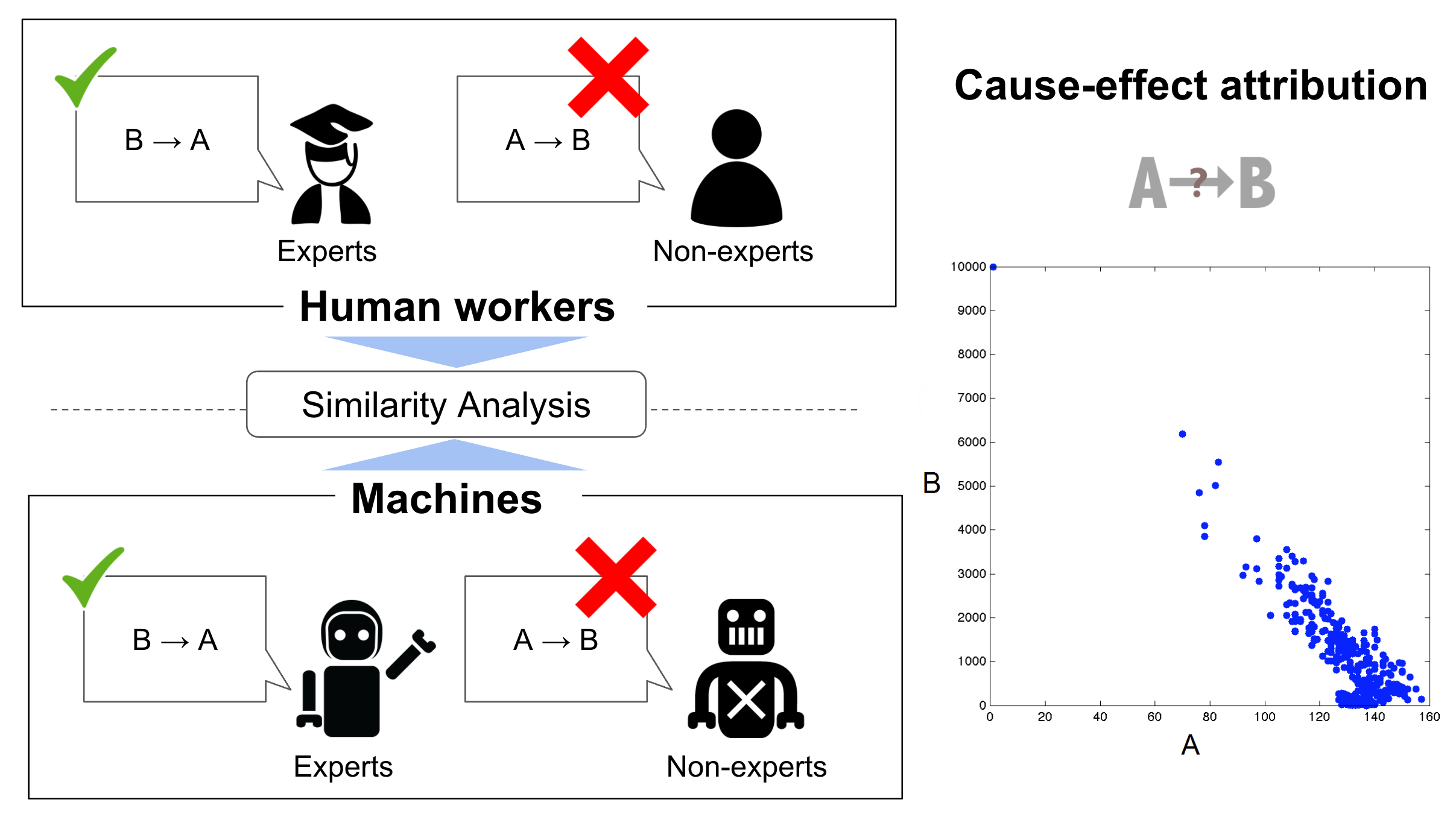}
    \caption{Summary of the paper.}
\end{figure}

The summary of this paper is shown in Figure \ref{fig:paper_summary}. In this study, we sought to understand how experts and non-experts solve a task's problems by comparing their performances with those of computational algorithms. Building a computational model that performs similarly to human experts or non-experts is a basic approach to understanding the difference in humans' cognitive abilities. Among computational models, we chose to use artificial neural networks (ANNs) because they have been widely used to model cognitive abilities on vision \cite{LeCun:1998ek} and languages \cite{Elman:1990wt}. ANNs were inspired by neurons in human brains and are considered powerful tools for building predictive models. Specifically, we used convolutional neural networks (CNNs) \cite{LeCun:1998ek} for our study. CNNs were inspired by human vision and originally developed for the handwritten digit-recognition task. Recent advances in CNN models have prevailed in a wide variety of fields, including image recognition, speech recognition, and natural language processing \cite{LeCun:2015dt}.


Despite the large number of studies on ANNs, it is still unclear how we can use ANNs to understand the differences between experts' and non-experts' cognitive abilities; therefore, the key challenge in this paper is how to differentiate human experts and non-experts using ANNs. Our research questions in this paper are as follows: ``What differentiates experts and non-experts in cause-effect attribution? (RQ1)'' and ``Can we provide a framework to understand the differences between experts and non-experts? (RQ2)''

This paper collected human judgments on cause-effect attribution problems and prepared predictive models based on the CNN technique to (1) evaluate the consistency between human and machine judgments and (2) understand experts' insights by analyzing the representations of the CNN model that perform similarly to human experts. 

Our contributions in this papers are as follows:
\begin{itemize}
  \setlength{\parskip}{0cm}
  \setlength{\itemsep}{0cm}
  \item We applied the deep learning technique to the cause-effect attribution task to model human cognitive abilities.
  \item We propose a framework that uses the CNN internal representations of the task to understand and evaluate expertise of the humans in this visually-represented task.
\end{itemize}

\section{Related Work}
Despite the long history of causality studies using machine-learning techniques, applying machine-learning techniques to infer causality based on observations of two variables has not been well studied. The Causality Challenge\footnote{\url{http://www.causality.inf.ethz.ch/cause-effect.php}}, run by the NIPS Workshop and the IJCNN Workshop, provides a data set of several thousands of variable pairs from a wide variety of domains. Researchers have conducted studies to try to automate causal judgments based on the following approaches: a) using causal inference techniques (i.e., probabilistic graphical models) to unravel cause-effect relationships \cite{brown2008strategy,Lopez-Paz:2015:RCC:2789272.2912092,saeed2008bernoulli} and b) developing discriminative models based on supervised machine learning and features directly extracted from cause-effect pairs by applying mathematical functions \cite{boulle2007compression,guyon2010causation,nikulin2009ensemble}. To the best of our knowledge, no researchers have compared the performances of machine algorithms and human workers to understand human judgments of causality.

Several studies have tried to understand the similarity and the difference between humans and machines including neural networks \cite{Lake:2016wm,Lake:2015ds,Dai:2017uw}. Lake et al. \cite{Lake:2016wm} stated that the difference between humans and neural networks can be summarized as two fold: (1) people learn from fewer examples. (2) people learn the concept of a task, which is more than how to recognize patterns. Following the previous studies, this paper aims to try to understand the performance of human workers through analyzing the similarity between humans and machines toward improving human workers. Also, our study is first to use the cause-effect attribution task to understand the similarity between humans and machines. We also distinguish experts and non-experts under the assumption that they have different cognitive abilities in the task.

Previous studies \cite{Donahue:2014ta,SimoSerra:2016cd,Veit:2015bk,Yu:2014wt} have tried to analyze internal representations of trained neural network models. Recent advances in analyzing deep neural networks allowed for the deep-layer application of manifold learning methods, such as $t$-SNE \cite{Maaten:2008tm}, to interpret high-level representations in an intuitive manner. Simo-Serra and Ishikawa \cite{SimoSerra:2016cd} and Veit et al. \cite{Veit:2015bk} used $t$-SNE to convert activations in the fully-connected layers of trained CNNs for grouping and recommending fashion styles based on the CNN models, and Johnson et al. \cite{Johnson:2016vk} used the same approach to cluster artistic styles of paintings. Donauhue et al. \cite{Donahue:2014ta} and Yu et al. \cite{Yu:2014wt} proposed a general framework of decoding trained neural network models into interpretable visualizations. This paper is the first to apply these methods of analyzing representations to the cause-effect attribution task for understanding human cognitive abilities to perform tasks based on visualizing internal representations of neural network models.

\section{Methodology}
The cause-effect attribution task involves classifying the cause-effect relationship between two variables based on a number of observations. Each pair belongs to one of three classes: (1) A causes B ({\tt forward}), (2) B causes A ({\tt backward}), or (3) No cause-effect relationship ({\tt no-causality}). We used the NIPS Workshop Causality Challenge dataset, which has been used for previous related studies \cite{guyon2010causation}\cite{Panov:2014ha}\cite{Mooij:2016uf}. The variable pairs that have causality relationships were selected from actual variables in a wide variety of fields, such as chemistry, ecology, economy, and so on; therefore, the pairs have ground-truth labels regarding their causality. The dataset consists of 4,050 pairs of variables. Each variable has numerical values. The dataset is described as $\{ ({\bf x}_A^{(i)}, {\bf x}_B^{(i)}, y^{(i)} ) \}_{i=1}^N$, for which $\mathbf{x}_A^{(i)} = (x_{A, 1}^{(i)}, x_{A, 2}^{(i)}, \dots, x_{A, n^{(i)}}^{(i)})^T$ are the values of variable A, $\mathbf{x}_B^{(i)} = (x_{B, 1}^{(i)}, x_{B, 2}^{(i)}, \dots, x_{B, n^{(i)}}^{(i)})^T$ are the values of variable B, $y^{(i)} \in \mathcal{Y}$ is the label of the $i$-th pair, $n^{(i)}$ is the number of values for the $i$-th pair, and $N$ is the number of pairs. $\mathcal{Y}$ is a set of causality relationships $\{1, -1, 0\}$ that correspond to {\tt forward}, {\tt backward}, and {\tt no-causality}, respectively. 

\subsection{Human Workers}
To collect human workers' judgments, we conducted crowd-sourced experiments at MTurk. We collected sixty annotators' judgments for sixty pairs that were randomly selected from the cause-effect dataset; each pair has sixty judgments by sixty annotators. Due to the fact that crowdsourcing workers are often not willing to accept tasks that take more than a certain length of time, we split the sixty pairs into five independent tasks to conduct the experiments efficiently.

\begin{figure}[t]
    \centering
    \subfigure[]{
        \includegraphics[scale=0.38]{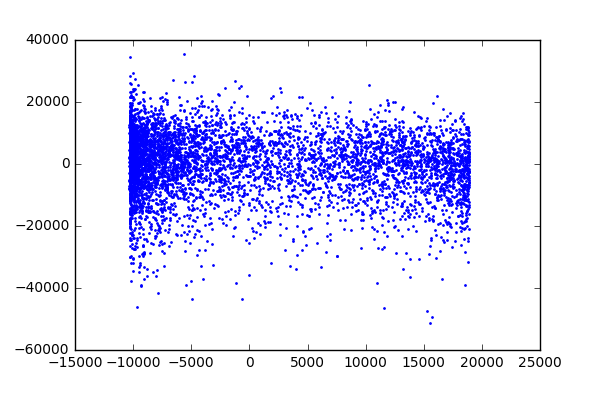} \label{fig:scatterplot-id11-label0}
    }
    \subfigure[]{
        \includegraphics[scale=0.38]{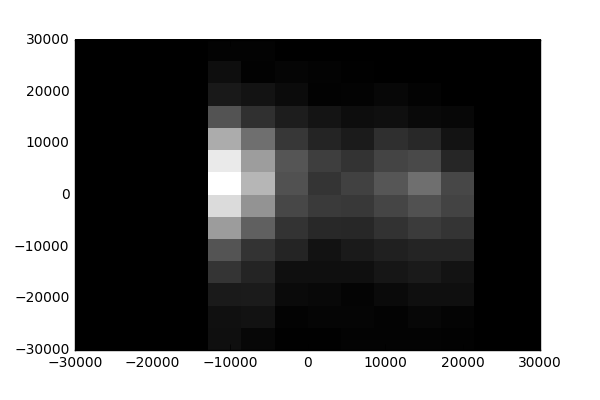} \label{fig:cnninput-id11-label0}
    }
\\
    \subfigure[]{
        \includegraphics[scale=0.38]{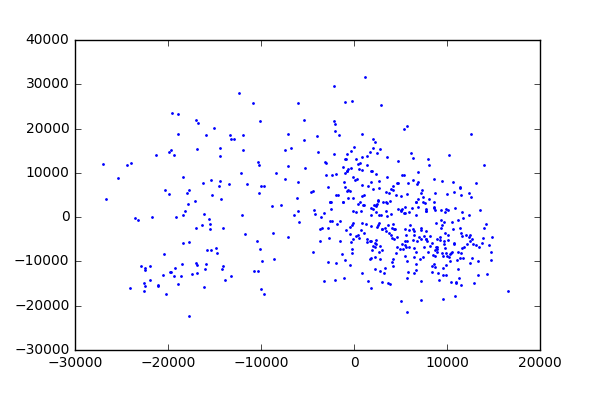} \label{fig:scatterplot-id83-label-1}
    }
    \subfigure[]{
        \includegraphics[scale=0.38]{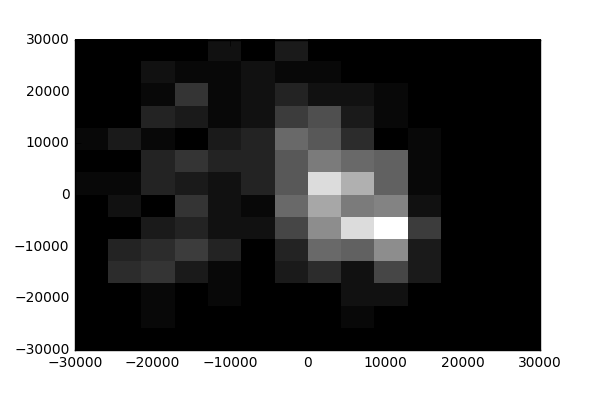} \label{fig:cnninput-id83-label-1}
    }
    \caption{(a) Scatter plot of the cause-effect attribution task sample (ID: 11, {\tt no-causality}). The scatter plot was shown to crowdsourcing workers to collect their judgments. (b) Coarse-grained visualization of the same sample as (a). The image was used as an input for the machine algorithms. The human expert group and expert machines' predictions were correct, while the predictions by the human non-expert group and non-expert machines were incorrect. (c) and (d) are the same figures for another sample (ID: 83, {\tt backward}). Only the prediction by expert machines was correct.}
\end{figure}

Each set of the tasks first presented {\it nine examples} (three for {\tt forward}, three for {\tt backward}, and three for {\tt no-causality}) in scatterplot images (e.g., \ref{fig:scatterplot-id11-label0} and \ref{fig:scatterplot-id83-label-1}) in the instruction and then asked the MTurk worker to answer cause-effect judgments for twelve images in a multiple-choice manner (``A causes B'', ``B causes A'', or ``None of them''.) The twelve images were randomly chosen from three classes so that the ratio of guessing the correct answer randomly were $1/3$.

For each set, we calculated the accuracy of the cause-effect attribution tasks performed by the sixty annotators. The annotators were sorted by their accuracy and divided into two groups according to their performances after low-performing annotators were filtered out. Specifically, we filtered out the annotators whose accuracy values were within the bottom one-third of annotators. This filtering was conducted to avoid the potential problem of unreliable submissions from crowdsourced workers. Then we split the remaining two-thirds of the annotators into {\it expert} and {\it non-expert} groups according to their accuracy values. These annotators' cause-effect attribution predictions were conducted in an aggregated manner for each group. That is, the majority vote of the submitted judgments for each cause-effect pair belonging to the expert group was used as the expert group prediction for that pair. 

\subsection{Machine Algorithms}
We also conducted an experiment to evaluate the algorithms' performances on cause-effect attribution. To maintain consistency with the human workers' experiments, we also used visual images of cause-effect pairs as features and the corresponding labels as target values in order to train a predictive model based on the machine learning technique. In this paper, we used the CNN \cite{LeCun:1998ek} as a machine learning algorithm because it is well-known for its high performance in image recognition and because it can also learn to interpret high-level representations of target domains.

An example of input images for the CNN is shown in Figure \ref{fig:cnninput-id11-label0}. The resolutions of the images were downgraded in order to maximize performance (high-resolution images caused overfitting problems). Note that we chose a resolution of 28x28, which performs best among other settings in the preliminary experiments. The other settings of CNN used in the experiments are described in Appendix.
 
We built multiple machine algorithms to prepare machine models that have different expertise in the cause-effect attribution task. In this paper, we assume that the {\it expertise} can be decomposed into two perspectives: (1) {\it cognitive ability} and (2) {\it knowledge}. That is, a machine algorithm with good cognitive ability to perform the cause-effect attribution task has a reasonably good function to convert raw images into reasonable representations for the task, while a machine algorithm with a good knowledge of the cause-effect attribution task is trained with a large amount of training data for better generalization. We modeled the two parameters, cognitive ability and knowledge, by (1) {\it representation learning quality} and (2) {\it number of training samples}. Two settings were prepared to differentiate each perspective; therefore, we used four machine algorithm methods in this experiment. Table \ref{table:machine_algorithms} summarizes the four machine algorithms' settings. 

In order to prepare an expert and a non-expert for the representation learning perspective, we used the cause-effect dataset ({\tt CE-*}), or the MNIST dataset ({\tt MNIST-*}), to train the CNN models. The MNIST dataset\footnote{\url{http://yann.lecun.com/exdb/mnist/}} consists of a large number of handwritten digits and ground-truth labels. We applied many neural network methods to the MNIST dataset because the handwritten digit-recognition task is considered to be a fundamental cognitive task. We assume that CNN models trained on the cause-effect dataset should be able to convert images of the cause-effect attribution task into reasonable representations while CNN models trained on the MNIST dataset should have reduced abilities in representation learning.

We prepared two settings for the knowledge perspective. The first setting used all the training data to train predictive models while the second setting used only nine samples to train predictive models. We considered the first setting a {\it knowledgeable} method; the second setting assumes the model does not have any prior knowledge on the cause-effect attribution task other than the nine examples shown to human workers with ground-truth labels during the MTurk experiments.

\begin{table}[t]
\centering
\caption{Four settings of machine algorithms. }\label{table:machine_algorithms}
\begin{tabular}{|c|c|c|} \hline
Methods          & Rep. learn. source & \# of train. data \\\hline
{\tt CE-all}      & CE       & 3,990    \\\hline
{\tt CE-9}    & CE          & 9  \\ \hline
{\tt MNIST-all}   & MNIST       & 3,990   \\ \hline
{\tt MNIST-9} & MNIST       & 9  \\ \hline
\end{tabular}
\end{table}

\subsection{Similarity Analysis}
To analyze the similarities between human workers' judgments and those of machine algorithms, we calculated the Pearson correlation between the human and algorithmic methods. The correlation coefficient was between two vectors of correct (as one) and incorrect (as zero) distributions; thus, the correlation value captures the tendency of the different methods' correct and incorrect predictions.

\subsection{Artificial Neural Representation Analysis}
This paper developed a framework for understanding human cognitive abilities. The framework used a trained CNN model as a representation extractor that converts the input representations of a sample into high-level representations. Precisely, the framework used $t$-SNE \cite{Maaten:2008tm} to embed the internal representations of the CNN model into 2-D space. The technique has been used in \cite{Donahue:2014ta,SimoSerra:2016cd,Veit:2015bk,Yu:2014wt} to qualitatively evaluate neural networks' learning. To the best of our knowledge, this is the first study that applies the neural representation technique to the understanding of human cognitive abilities in performing the cause-effect attribution task.

\section{Results and Discussion}
{\bf Human experts vs. expert machines.} First, we compared the performances of human experts and two ``expert'' machines (i.e., {\tt CE-all} and {\tt CE-9}) which should have reasonable internal representations for the cause-effect attribution task. The accuracy values of human experts, {\tt CE-all}, and {\tt CE-9} were 0.600, 0.650, and 0.450 respectively. The results show that human experts outperform the expert machine that knows only a few examples ({\tt CE-9}) while human experts do not perform as well as the ``knowledgeable'' expert machine ({\tt CE-all}.) The correlation coefficient between human experts and {\tt CE-all} was 0.328 ($p < 0.05$) and the correlation coefficient between human experts and {\tt CE-9} was 0.260 ($p < 0.05$). These correlation values are much higher than those with the non-expert humans (or the non-expert machines, in the next paragraph). Thus, the results support our conjecture that human experts have better internal representations of the cause-effect attribution task than human non-experts.

{\bf Human non-experts vs. non-expert machines.} Second, we compared the performance of human non-experts and two ``non-expert'' machines (i.e., {\tt MNIST-all} and {\tt MNIST-9}) which should have poor internal representations for the cause-effect attribution task because they were not trained on the specific task. The accuracy values of human non-experts, {\tt MNIST-all}, and {\tt MNIST-9} were 0.433, 0.467, and 0.500 respectively. The results show that human non-experts are worse than the non-expert machines. The correlation coefficient between human non-experts and {\tt MNIST-all} was 0.058 (no statistical significance) and the correlation coefficient between human non-experts and {\tt MNIST-9} was 0.269 ($p < 0.05$.) The results support the conjecture that human non-experts have an internal representation that is only somewhat appropriate for the task and also have a weak model of the task.

{\bf All results.} All of the methods' evaluation results are shown in Table \ref{tab:results}. Table \ref{tab:correlation} summarizes the correlation values between machine algorithms and human experts and non-experts. 
 
\begin{table}[t]
\centering
\caption{Accuracy of methods for sixty test samples.}\label{tab:results}
\begin{tabular}{|c|c|c|}
\hline
Type    & Model     & Accuracy \\ \hline
Human   & Expert   & 0.600    \\ \hline
             & Non-expert   & 0.433    \\ \hline
Machine    & {\tt CE-all} & 0.650    \\ \hline
                & {\tt CE-9} & 0.450    \\ \hline
                & {\tt MNIST-all} & 0.467    \\ \hline
               & {\tt MNIST-9}   & 0.500    \\ \hline
\end{tabular}
\end{table}

\begin{table}[t]
\centering
\caption{Pearson correlation values between human methods and machine methods. * denotes $p < 0.05$ and bold face denote the highest correlation value in a column.} \label{tab:correlation}
\begin{tabular}{|c|c|c|}\hline
Methods          & Human expert       & Human non-expert \\\hline
CE-all    & ${\bf 0.328}^*$          & 0.148            \\\hline
CE-9      & $0.260^*$          & 0.020            \\\hline
MNIST-all & 0.218              & 0.058 \\\hline
MNIST-9   & 0.204              & ${\bf 0.269}^*$  \\\hline
\end{tabular}
\end{table}

\begin{figure*}[t]
    \centering
    \subfigure[]{
        \includegraphics[width=0.46\textwidth]{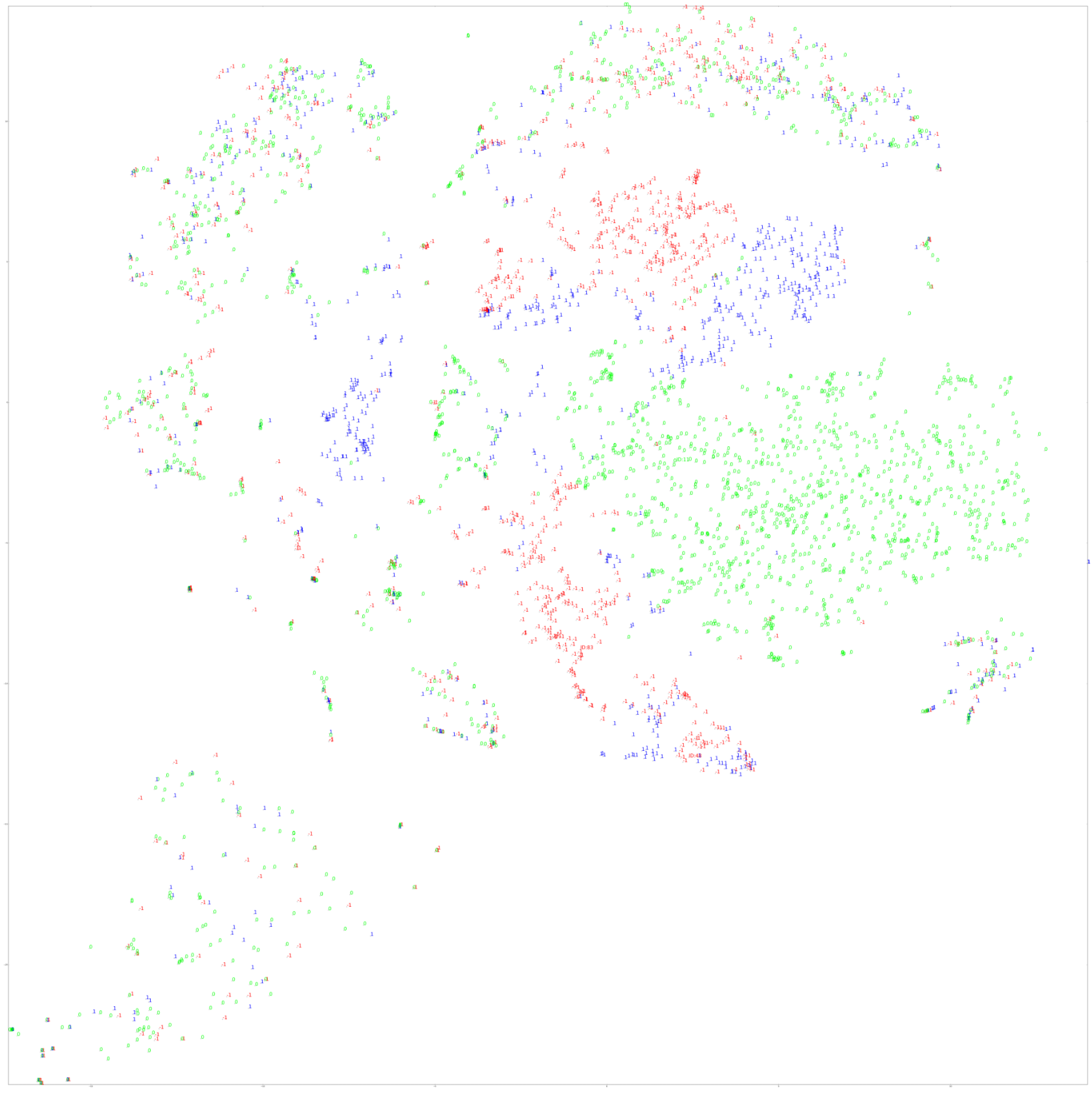} \label{fig:ce-tsne-all}
    }
    \subfigure[]{
        \includegraphics[width=0.46\textwidth]{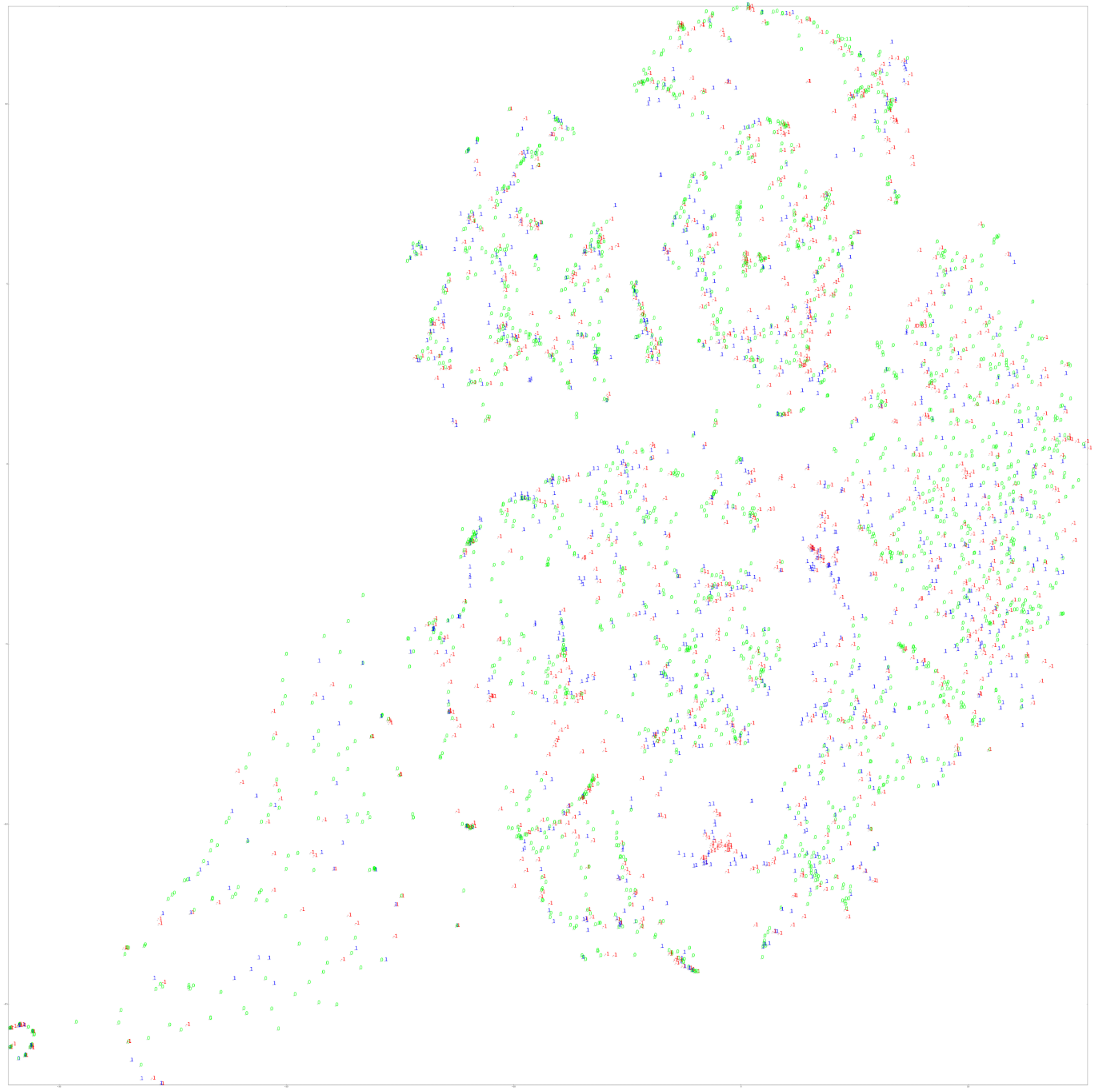} \label{fig:mnist-tsne-all}
    }
    \subfigure[]{
        \includegraphics[width=0.46\textwidth]{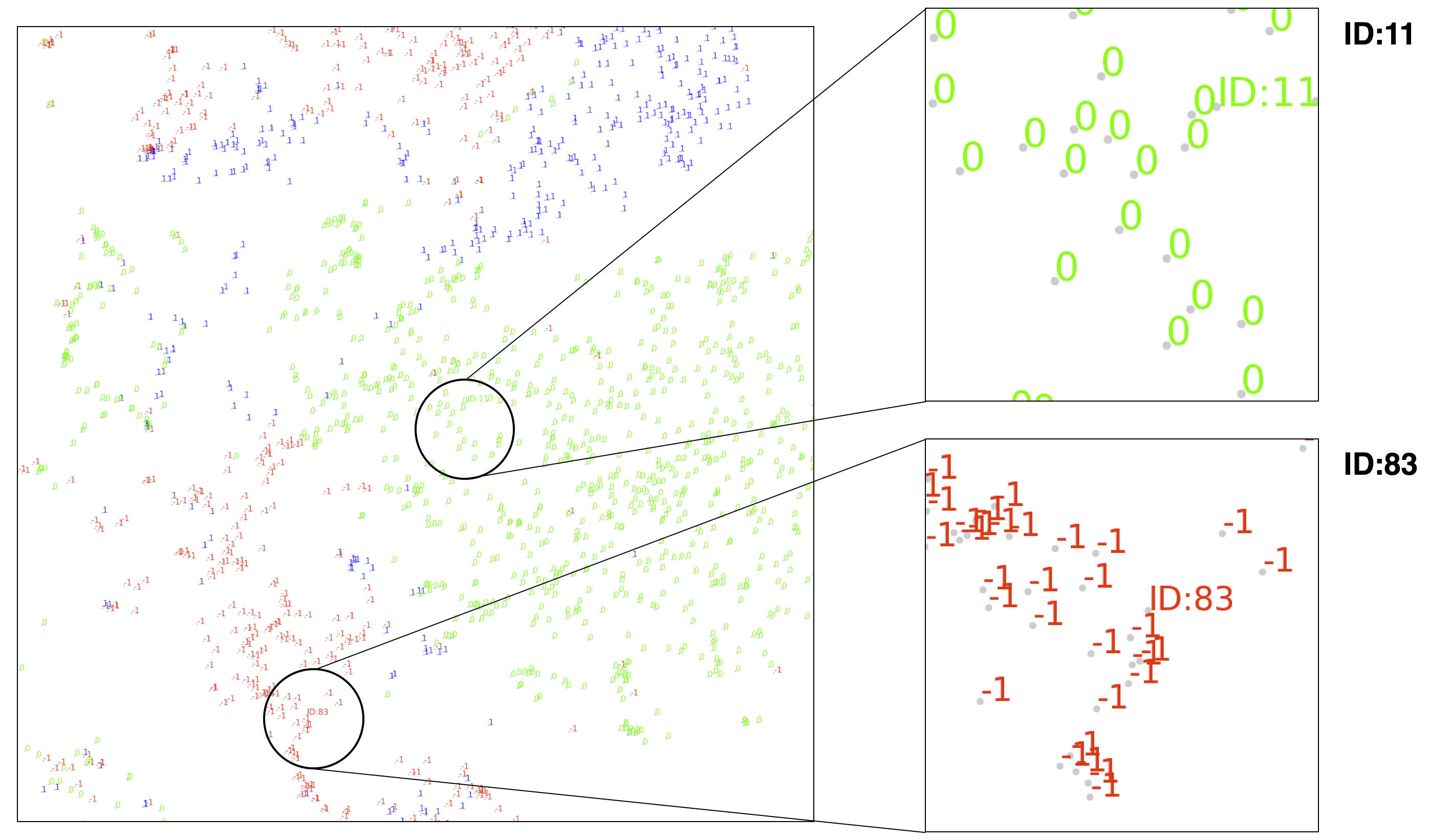} \label{fig:ce-tsne}
    }
    \subfigure[]{
        \includegraphics[width=0.46\textwidth]{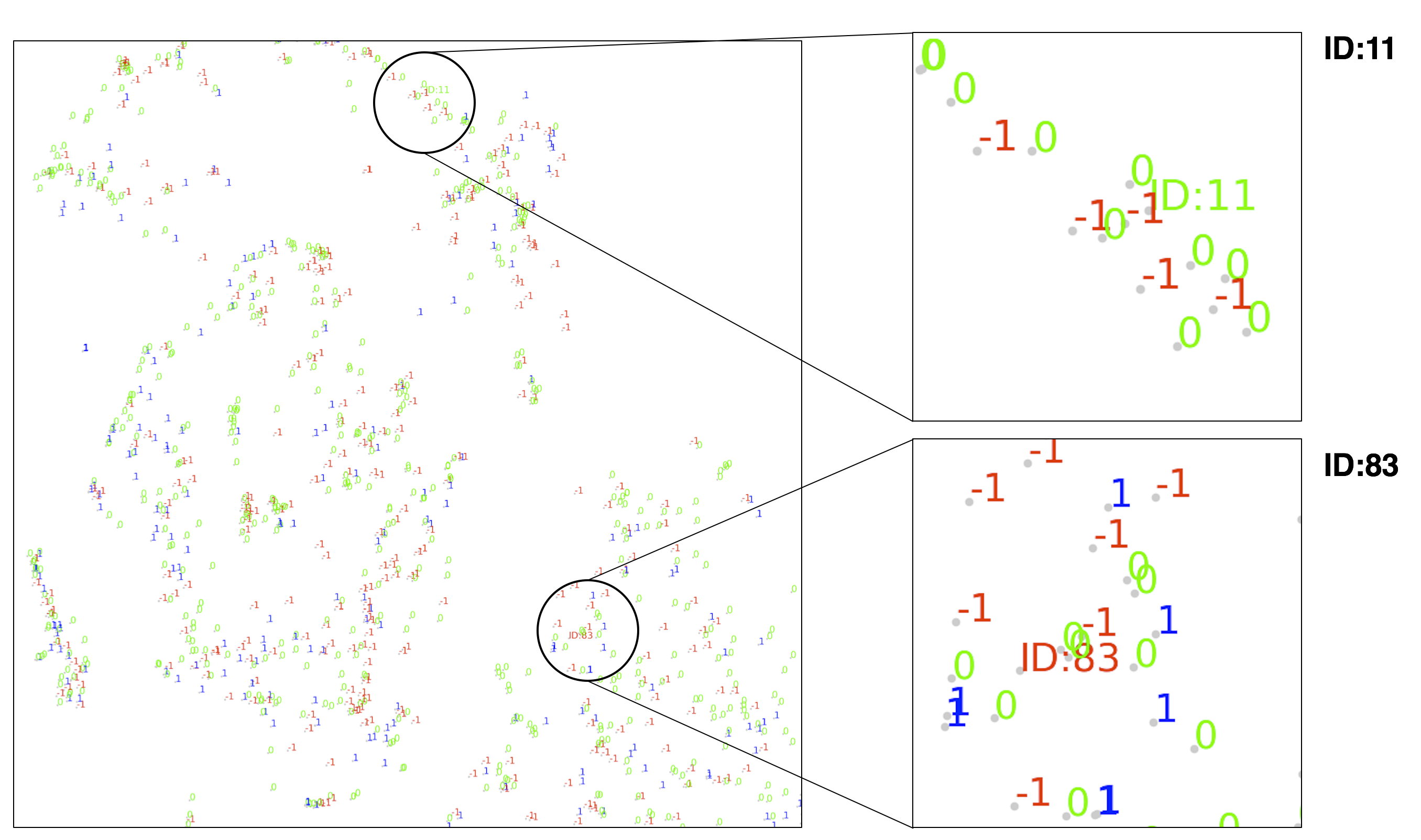} \label{fig:mnist-tsne}
    }
    \caption{(a) Artificial neural representation of the {\tt CE-*} and (b) Artificial neural representation of {\tt MNIST-*}: Each point corresponds to the sample in the cause-effect dataset. The symbol 0 denotes {\tt no causality}, 1 denotes {\tt forward}, and -1 denotes {\tt backward}. (c) The {\tt CE-*} and (d) {\tt MNIST-*} show the magnified region around sample ID 11 and sample ID 83.}
\end{figure*}

{\bf Neural Representation Analysis.}
To understand representation learning, we analyzed {\it neural representations} of {\tt CE-*} and {\tt MNIST-*} by investigating the high-level representations. We used $t$-SNE to convert 128-dimensional vectors into 2-D vectors to visually map the cause-effect dataset samples into 2-D scatter plots. Figures \ref{fig:ce-tsne} and \ref{fig:mnist-tsne} show the 2-D scatter plots of {\tt CE-*} and {\tt MNIST-*}. The points in the scatter plots are the mapped samples of the cause-effect dataset; the symbols and colors denote the corresponding labels (1 (blue): {\tt forward}, -1 (red): {\tt backward}, 0 (green): {\tt no-causality}.) The magnified regions of the 2-D scatter plots (Figure \ref{fig:ce-tsne} and \ref{fig:mnist-tsne}) show the surrounding spaces of the samples (ID:11 and ID:83) in the cause-effect dataset.

\begin{table*}[t]
\centering
\caption{Network architecture of CNN. Parentheses denote the activation function of the layer. FC denotes a fully-connected layer. The DropOut technique was applied to the output of the fourth and the fifth layers.}\label{tab:network_architecture}
\small
\begin{tabular}{|c|c|c|c|c|c|c|}\hline
Index & 1 & 2 & 3 & 4 & 5 & 6 \\\hline
Name & Input & Conv (ReLU) & Conv (ReLU) & Max-pooling & FC (ReLU) & Output (softmax) \\\hline
Filter \# & 1 & 32 & 32 & 32 & 128 & 3 \\\hline
\end{tabular}
\normalsize
\end{table*}

\section{Summary}
We conducted an empirical study for understanding human cognitive abilities to attribute cause-effect relationships based on visual information. Our study compared the performances of human workers and machine algorithms in order to understand human experts' and non-experts' cognitive abilities to perform the task. The comparison showed that the majority vote of human experts had a significant correlation with a machine algorithm that used a CNN trained on the target domain for feature extraction and that used the entire training dataset for classification. The analysis also showed that non-experts humans' judgments had a significant similarity to a machine algorithm that used a non-target domain dataset to train the CNN model for feature extraction and then used only a limited amount of training data for classification. 

We also developed a framework for qualitatively analyzing the accuracy of neural network models' internal representations. Our method mapped data into 2-D spaces based on the activations of nodes in a fully connected layer of a CNN model. The method enabled us to interpret how a CNN model's representation learning creates clusters of data in a target domain. Our framework analysis showed that human experts successfully converted original representations into appropriate internal representations that created reasonable clusters in the 2-D space, while human non-experts rarely converted original representations into high-level representations.

Our study contributes to the field of cognitive computing by taking a step toward understanding human cognitive abilities to perform the cause-effect attribution task, and our framework provides a novel way to understand human cognitive abilities in general. We believe that this study will contribute to the development of systems that take into account human expertise in facilitating the education of human workers and supporting the collaboration between human workers and machine algorithms.

\bibliographystyle{unsrt}
\bibliography{refs}

\appendix
\section{Appendix}
\subsection{CNN Settings}
The network architecture of the CNN models used for the experiment is shown in Table \ref{tab:network_architecture}. The CNN prepared two convolution filters, followed by the max pooling layer with DropOut \cite{Srivastava:2014ww} and the fully-connected layers. Each layer, except for the input, max-pooling, and output layers, used ReLU as its activation functions. The method basically follows the network architecture of LeNet \cite{LeCun:1998ek}. The filter size of these convolution filters was 4x4; there were 32 feature maps, and there were 128 nodes in the fully-connected layer. ADADELTA \cite{Zeiler:2012uw} was used as an optimization technique. The batch size was 128 and the number of epochs was 359. These network parameters were chosen after preliminary experiments so that the CNN could perform the cause-effect attribution task as efficiently as possible. For {\tt CE-*}, the CNN model was trained on the 3,990 training examples that excluded the 60 test examples used for the MTurk experiments. For {\tt MNIST-*}, the CNN model was trained on the 60,000 samples of the training sets of the MNIST dataset. Note that the method also used 28x28 input representations for consistency with the cause-effect attribution task and changed the size of the CNN's 6th (output) layer used for {\tt CE-*} from three to ten to fit the number of the MNIST task's classes.

\subsection{Classification}
After fitting a CNN model with training data, we were able to extract a high-level representation of a sample based on the CNN model. The method input each sample as a 28x28 representation and obtained the activated values in the fully connected layer (the fifth layer) via forward propagation. As a result, each sample was converted into a 128-dimensional vector representation. The vector representations can be considered high-level representations obtained from {\it cognitive abilities} of {\tt CE-*} or {\tt MNIST-*}. In this paper, we used the $k$-NN algorithm for classification. For instance, {\tt CE-all} finds the $k$-nearest neighbor(s) of an input sample from the whole of the training dataset (i.e., 3,990 samples) to classify the sample, while {\tt CE-9} finds the $k$-nearest neighbor for classification in the same manner from the nine samples rather than all 3,990 samples. To find $k$-nearest neighbors, we calculated the distance based on Euclidean distance in the 128-dimensional space. Note that we used $k$-NN instead of fine-tuning the CNN models' entire network architecture because we wanted to differentiate the performances of representation learning by different datasets. Also, our aim was not to achieve high accuracy in the task but to understand human cognitive abilities through analyzing neural network models.

\end{document}